\documentclass[letterpaper]{article} 
\usepackage{aaai24}  
\usepackage{times}  
\usepackage{helvet}  
\usepackage{courier}  
\usepackage[hyphens]{url}  
\usepackage{graphicx} 
\usepackage{natbib}  
\usepackage{caption} 
\usepackage{algorithm}
\usepackage{algorithmic}

\usepackage{newfloat}
\usepackage{listings}
\usepackage{multirow}
\usepackage{amsmath}
\usepackage{amssymb}

\DeclareCaptionStyle{ruled}{labelfont=normalfont,labelsep=colon,strut=off} 
\floatstyle{ruled}
\newfloat{listing}{tb}{lst}{}
\floatname{listing}{Listing}
\title{NegVSR: Augmenting Negatives for Generalized Noise Modeling in Real-world Video Super-Resolution}
\author{
    Yexing Song\textsuperscript{\rm 1},
    Meilin Wang\textsuperscript{\rm 1},
    Zhijing Yang\textsuperscript{\rm 1},
    Xiaoyu Xian\textsuperscript{\rm 2},
    Yukai Shi\textsuperscript{\rm 1}\thanks{Corresponding author.}
}
\affiliations{
    \textsuperscript{\rm 1}School of Information Engineering, Guangdong University of Technology\\

    \textsuperscript{\rm 2}CRRC Academy 

    Songsheng0505@gmail.com, xxy@crrc.tech, \{wml, yzhj, ykshi\}@gdut.edu.cn
}

\usepackage{bibentry}

\begin{document}

\maketitle

\begin{abstract}
The capability of video super-resolution (VSR) to synthesize high-resolution (HR) video from ideal datasets has been demonstrated in many works. However, applying the VSR model to real-world video with unknown and complex degradation remains a challenging task. First, existing degradation metrics in most VSR methods are not able to effectively simulate real-world noise and blur. On the contrary, simple combinations of classical degradation are used for real-world noise modeling, which led to the VSR model often being violated by out-of-distribution noise. Second, many SR models focus on noise simulation and transfer. Nevertheless, the sampled noise is monotonous and limited. To address the aforementioned problems, we propose a Negatives augmentation strategy for generalized noise modeling in Video Super-Resolution (NegVSR) task. Specifically, we first propose sequential noise generation toward real-world data to extract practical noise sequences. Then, the degeneration domain is widely expanded by negative augmentation to build up various yet challenging real-world noise sets. We further propose the augmented negative guidance loss to learn robust features among augmented negatives effectively. Extensive experiments on real-world datasets (e.g., VideoLQ and FLIR) show that our method outperforms state-of-the-art methods with clear margins, especially in visual quality. Project page is available at: \url{https://negvsr.github.io/}.
\end{abstract}

\section{Introduction}

Video super-resolution (VSR) is the process of changing from low-resolution (LR) video to high-resolution (HR) video. Currently, VSR is divided into traditional VSR and real-world VSR~\cite{RealBasicVSR}, depending on the existence of the HR labels. Nevertheless, the VSR model frequently suffers from overfitting to a specific dataset within a fixed domain, which leads to the test results are often violated by unknown degeneration~\cite{RealSR}. Due to the domain gap, traditional VSR methods often fail to reconstruct real-world images effectively. Thus, it is crucial to develop a more robust restoration system for VSR.

The primary objective in the real-world VSR task is to extract more representative spatial structures and reasonable texture details from images. Many works~\cite{RealSR, Real_ESRGAN, shi2020ddet, wei2020component} have ensured that the real-world model can produce high-quality images across various domains. For instance, Real-ESRGAN~\cite{Real_ESRGAN} proposed a high-order degradation model that better simulates real-world degradation. They expand the degeneration domain by a second-order degeneration model composing various classical degeneration kernels. But the high-order degradation mode has a theoretical upper bound on the degradation domain, which means the permutations of all the classical degenerate kernels are included. However, this strategy solely deals with a limited portion of real-world scene degradation.

\begin{table}[]
\centering
\begin{tabular}{l|c}
\textbf{Methods}           & \textbf{NIQE} $\downarrow$  \\ \hline
Mixup~\cite{mixup}             & 3.635 \\ \hline
CutOut~\cite{cutout}            & 3.563 \\ \hline
CutMix~\cite{cutmix}            & 3.470 \\ \hline
FMix~\cite{fmix}              & 3.585 \\ \hline
Mixup Noise       & 3.643 \\ \hline
NegMix, w/ $\mathcal L_{Aug-NP}$  (ours full)  & \textbf{3.188}
\end{tabular}
\caption{The quantitative comparison of our model with various data augmentation methods uses REDS and FLIR training dataset as the training video. 'Mixup Noise' denotes the mixing of VSR inputs with a solitary noise extracted from the FLIR training dataset. 'NegMix' is our negative augmentation method for VSR. The performance evaluation is performed on the VideoLQ dataset.}
\label{tab:mixups}
\end{table}

\begin{figure*}[h]
\centering
\includegraphics[width=0.98\linewidth]{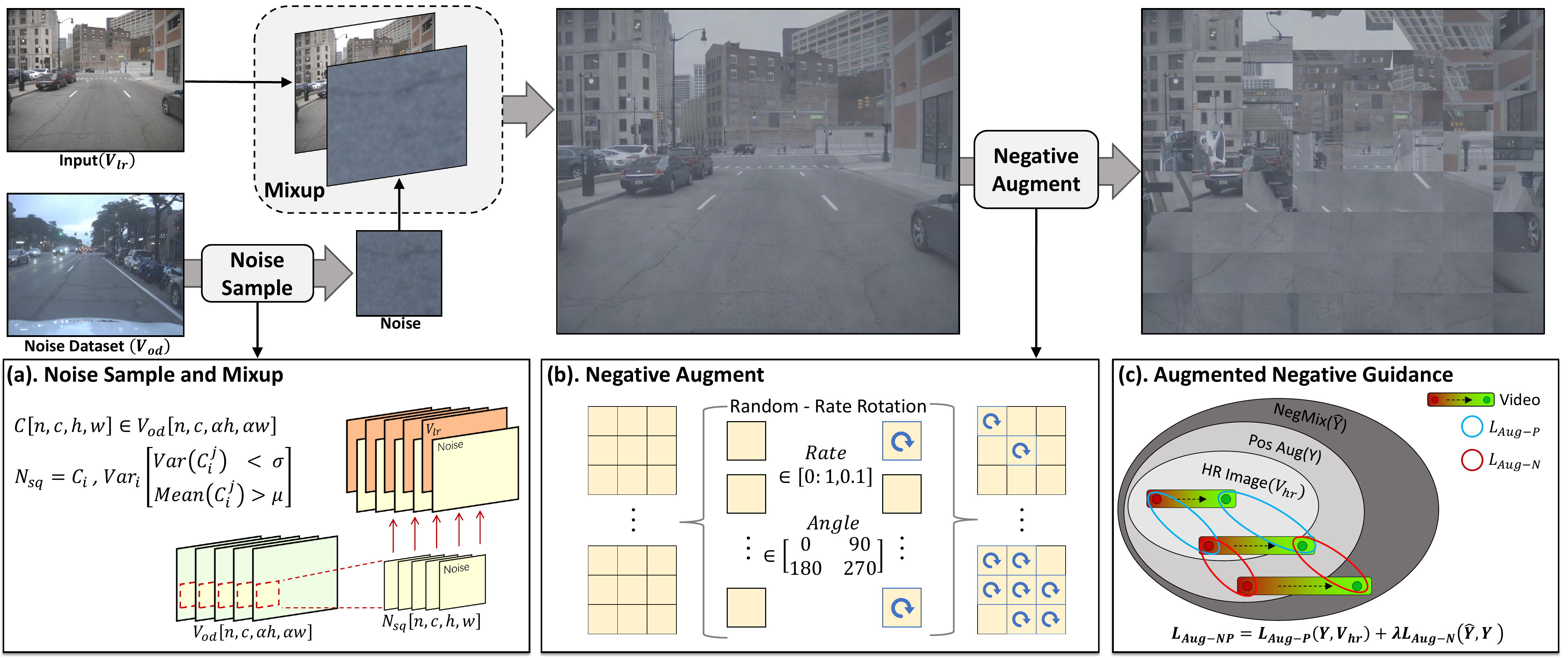}
\caption{The overview of the proposed NegVSR. (a) Our approach initially extracts noise sequence $N_{sq}$ through window sequence $C$ in an unsupervised manner. The motion of $C$ occurs within the OOD video noise dataset $V_{od}$. Subsequently, it mixes $N_{sq}$ and LR video $V_{lr}$ to create novel training input $V_{lr}^N$. (b) $V_{lr}^N$ is applied with a patch-based random central rotation to derive $V_{neg}$. (c) Both $V_{neg}$ and $V_{lr}$ are fed into the VSR model to generate $\widehat{Y}$ and $Y$, respectively. And $\mathcal{L}_{Aug-P}$ enables the model to recover realistic pixels from the $V_{lr}$. $\mathcal{L}_{Aug-N}$ drives $Y$ to learn the robust features present in the negative output $\widehat{Y}$.}
\label{fig:farmework}
\end{figure*}

Recently, many noise migration and simulation methods~\cite{RealSR, denoise_noisedataset_guidence,dong2023deep,pan2023deep} can extract the noise from the real-world dataset. They sample noise by calculating the feature from the real-world scene dataset. Estimating blur kernels and noise by modeling real-world noise effectively improves the quality of reconstructed images~\cite{BSRGAN}. Furthermore, suppose the sampled noise is mixed with the VSR input during training. The high-level semantic information in the input image will be further degraded, which helps the discriminative model learn robust features. However, in the VSR task, the noise domain shows a different pattern with space-time structure in the same video sequence, leading to misaligned information in the space-time dimensions. It reveals that the concept of concurrently processing sequential frames and independent noise needs to be re-examined. As illustrated in Tab.~\ref{tab:mixups}, 'Mixup Noise' comparisons with other mixing methods produce the worst result. Therefore, one of the primary challenges in real-world VSR is to investigate sequential noise sampling algorithms corresponding to the space-time dimension in the video sequence.

In this paper, we develop a sequential noise modeling approach for the real-world VSR. The proposed method consists of three main stages: noise sequence sampling, negative sample/noise augmentation, and recovery via augmented negative guidance. First, our approach samples noise sequences in an unsupervised manner from the out-of-distribution (OOD) video noise dataset $V_{od}$ and mixes the noise sequence with the training video. Meanwhile, the sampled noise sequence contains information in both the temporal and spatial dimensions, which will allow the VSR model to learn high-order degradation among real-world noise sequences. Second, we propose a negative augmentation for video frames and sequential noise. Specifically, we perform a patch-based center rotation operation on the video. The proposed negative augmentation operation preserves the semantic information of the local region but destroys the spatial connections between patches, reducing global semantic information, which creates a more challenging degradation metric. Finally, we propose the augmented negative guidance loss to effectively learn robust features among augmented negatives. To demonstrate the effectiveness of our proposed approach, we conduct experiments on two real-world video datasets: VideoLQ~\cite{RealBasicVSR} and FLIR. In both datasets, our approach achieved superior performance in terms of quantitative and qualitative indexes. Additionally, we perform an ablation study to evaluate the effectiveness of each component in our method.

In summary, our overall contributions are summarized in four-fold:
\begin{itemize}
    \item We re-examine the traditional noise mixup strategy in the VSR task and introduce a video noise sampling method that can extract the noise sequence from a given video in an unsupervised manner while ensuring that the space-time information within the noise sequence is continuous. 
    \item We propose a negative augmentation for generalized no-\\ise modeling. With the negative augmentation, NegVSR aims to create various yet challenging sets of real-world noise.
    \item We employ an Augment Negative Guidance loss to learn robust features from augmented negatives and enhance model generalization ability. 
    \item Our extensive experiments on two real-world datasets demonstrate that NegVSR outperformed not only other advanced methods but is also highly effective in noise reduction.
\end{itemize}

\section{Related Work}
\textbf{Video Super-Resolution. }VSR is an extension of SISR (Single-Image Super-Resolution)~\cite{sisr_2}. Unlike SISR, VSR necessitates the utilization of information contained in multiple frames. Existing VSR research~\cite{EDVR_location_frame_information} points out that effectively utilizing the information contained in frames can improve the performance of VSR. And the alignment module is commonly utilized to leverage inter-frame information. VSR methods using alignment module can be categorized into two groups: estimation and compensation~\cite{BasicVSR, TecoGAN_location_MEMC} and dynamic convolution (DCN)~\cite{TDAN_location_DC,  BasicVSR++}. Recently, BasicVSR~\cite{BasicVSR} introduced a bidirectional propagation module aggregating information from future and past frames. BasicVSR++~\cite{BasicVSR++} builds upon BasicVSR by incorporating additional backward and forward propagation branches. Furthermore, BasicVSR++ introduces optical flow alignment and DCN alignment, where optical flow alignment assists DCN alignment in achieving better performance.

\textbf{Real-World Video Super-Resolution. }Recent works in real-world VSR have focused on obtaining a larger unknown degeneration domain. RealVSR~\cite{RealVSR} utilizes a dual-lens phone camera to acquire LR-HR video pairs. 
Real-ESRGAN~\cite{Real_ESRGAN} incorporates a high-order degeneration model based on classic degeneration kernel combinations. AnimeSR~\cite{AnimeSR} employs convolution layers between degradation kernels. Nonetheless, expanding the domain of degeneration gives rise to the challenge of restoring high-quality video from a more complex degradation space. To tackle this problem,  RealBasicVSR~\cite{RealBasicVSR} introduces a dynamic cleaning module that suppresses degradation. FastRealVSR~\cite{FastRealVSR} proposes manipulating the hidden states to reduce artifacts.

\textbf{Noise Modeling. }Noise modeling has been utilized in many recent SR tasks. RealSR~\cite{RealSR} extracts noise by calculating the variance and injects noise into the input. GCBD~\cite{denoise_GAN} trains a Generative Adversarial Network (GAN) to estimate the noise distribution of the input noise and generate noise samples. RWSR-EDL~\cite{denoise_noisedataset_guidence} introduces a Noise-Guidance Data Collection method to address the time-consuming training required for optimizing multiple datasets. Our work presents the first proposal to utilize real-world noise sequence modeling in real-world VSR to enhance the network denoising capability. 

\section{Method}
In this section, we provide a detailed description of negative augmentation in NegVSR. First, we discuss the characteristics and challenges associated with the Mixup family. Second, we present a real-world noise sequence sampling and negative modeling method for VSR. The real-world noise sequence used for mixing is extracted unsupervised, but simple input-noise pair mixing methods can often lead to missing details. Finally, to address this problem, we propose a negative augmented noise-guided modeling approach. Through negative augmentation, VSR improves the ability to denoise robustly. During training, the LR video dimension $V_{lr}\in\mathbb R^{n \times c  \times h \times w}$  is equal to the real-world noise sequence $N_{sq}\in\mathbb R^{n \times c  \times h \times w}$. $h=w=64$ represents the size of the training input. 
\subsection{Preliminaries}
\label{sec:preliminaries}
\textbf{Mixup}~\cite{mixup} is a data augmentation methodology frequently employed in deep learning to enhance the model generalization capability. It produces novel training instances via a weighted amalgamation of pre-existing examples and their corresponding labels. Specifically, an additional sample is chosen randomly from the training dataset. And then, the two examples are combined convexly to construct a new example in both the input and label space. Mixup can be formulated as: 

\begin{equation}
\widetilde{x} = M \cdot  x_i +(1-M ) \cdot x_j,
\end{equation}
\begin{equation}
\widetilde{y} = M  \cdot y_i +(1-M ) \cdot y_j,
\end{equation}
where $x_{i}$ and $x_{j}$ represent the training samples, $y_{i}$ and $y_{j}$ denote their respective labels. $\widetilde{x}$ and $\widetilde{y}$ correspond to the new input and label. $M  \in  [0, 1]$ is the hyperparameter used in the Mixup. 

Mixup has inspired a range of variants and derivatives, which are demonstrated comprehensively in Tab.~\ref{tab:mixups}.

\begin{figure}[t]
\centering
\includegraphics[width=0.98\linewidth]{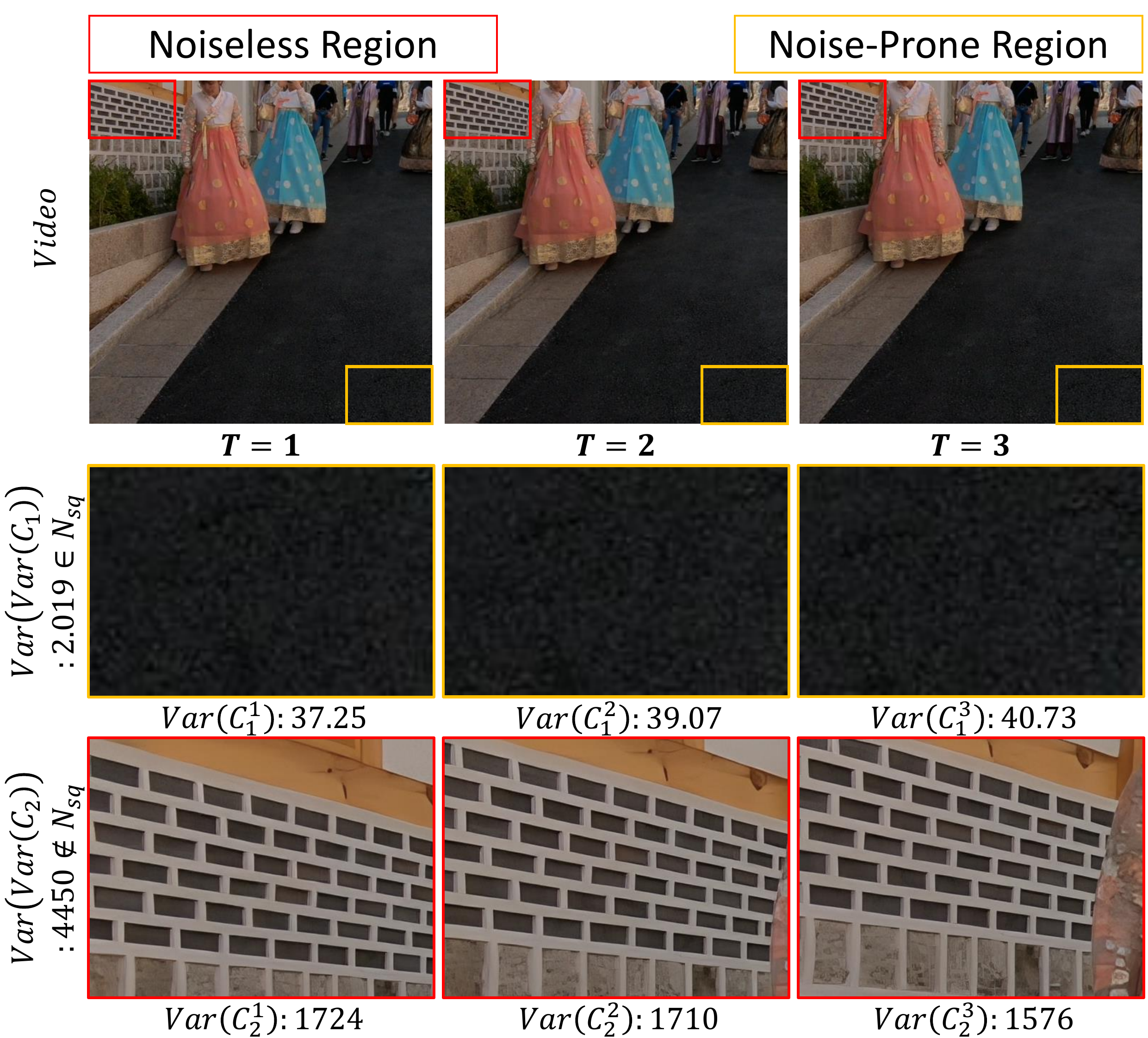}
\caption{Two window sequences $C_1$ and $C_2$ originate from the same video, which comprises consecutive frames. Based on this sliding window sequence strategy, the sequential Noise-Prone Region (yellow box) that contains less texture and more noise is selected by the low variance feature for noise augmentation. }
\label{fig:NoiseSequence}
\end{figure}

\subsection{Negatives for Generalized Noise Modeling}
\label{sec:negative_generation}
Real-world VSR differs from non-blind VSR due to the absence of labels. Specifically, non-blind VSR fails to resolve the various disruptions and changes in external video. This deficiency often results in worse performance in the OOD case. 

Considering the deficiency of non-blind VSR, we attempt to investigate a practical noise sampling strategy for sequential real-world frames. Assuming that the LR image $I_{lr}$ is degraded from the HR image $I_{hr}$, the single image degradation formula can be described as:
\begin{equation}
I_{lr}=(I_{hr}*K)_{\downarrow} + N,
\end{equation}
where $K$ is the blur kernel, and $N$ is the noise. 

In the video, most sequential data are taken from the same device and have a similar noise distribution. It implies that the noise domain has a strong connection across most frames within a sequence for video. Current noise modeling methods involve independent noise sampling within the input image and transfer those noise to more data for augmentation. However, it is only applicable to the single-image condition. It is essential for a video to ensure that the sampled noise remains consistent across frames within a sequence and keeps independent and identically distributed (i.i.d) across the different sequences. Therefore, we first propose a sequential noise sample and negative augmentation strategy for real-world video.

\textbf{Sequential Real-world Noise Generation.}
Building up-\\on the aforementioned observation, we present our proposed method for extracting sequential noise in video. As shown in Fig.~\ref{fig:farmework} (a). Suppose the video $V=[I_1I_2 \dots I_n] \in \mathbb{R}^{n \times c \times ah \times aw}$ contains $n$ frames and the image at moment $t$ is $I_t  \in \mathbb{R}^{ c \times ah \times aw}$. We scan the entire video using the window sequence $C$, each with a dimension of $\mathbb{R}^{n \times c \times h \times w}$. The total number of window sequences in a $V$ is $\frac { ah \times aw } { h \times w }=a^2$. The window of a window sequence $C_i$ at the $j$ moment is denoted by $C_{i}^j \in \mathbb{R}^{ c \times h \times w}$. Each window sequence $C_{i}$ contains $n$ windows $C_i=[C_i^1 C_i^2  \dots C_i^n]$. As is shown in Fig.~\ref{fig:NoiseSequence}. We calculate the variance for $C_{i}^{j}$. The scan window with high variance typically contains rich textures. These textures can impact the model to learn the noise distribution. High-variance window is commonly referred to as the noiseless region. Conversely, the noise in the window with low variance is perceptible. This window is referred to as the noise-prone region~\cite {decoupled_Mix}.

To ensure that the texture and margin in the extracted noise sequence are as uniform as possible~\cite{RealBasicVSR}, we need to calculate the variance for the mean and variance of each window in the sequence as follows:

\begin{equation}
    Var_i[Var(C_{i}^j)<\sigma] \in [0, \sigma_{var}],
\label{eq:var_var}
\end{equation}
\begin{equation}
    Var_i[mean(C_{i}^j)>\mu] \in [0, \sigma_{mean}],
\label{eq:var_mean}
\end{equation}
where $Var( \cdot )$ and $Mean( \cdot )$ refer to the functions used to calculate variance and mean, respectively. $\sigma$ and $\mu$ are the mean and variance of each window $C_{i}^j $. $\sigma_{var}$ and $\sigma_{mean}$ are the variance of the variance and mean of the window sequence $C_{i} $. We consider the window sequence $C$ that satisfies the Equ.~\ref{eq:var_var} and Equ.~\ref{eq:var_mean} as real-world noise sequence $N_{sq}$. Before training, we collect all the $N_{sq}$ to create an offline noise dataset. 

\textbf{Video Negative Augment for Generalized Noise Generation.}
\label{sec:Video Negative Augment for Generalized Noise Generation} 
We first extract $N_{sq}$ from $V_{od}$ and then mix $N_{sq}$ with $V_{lr}$ to generate the new training sample as follows:
\begin{equation}
    V_{lr}^N = M \cdot N_{sq}  +(1-M )\cdot V_{lr},
\end{equation}
where $M  \in  [0, 1]$ denotes the mixing noise weight. $M=1$ represents the new training input consisting entirely of $N_{sq}$.

VSR can effectively learn to denoise by incorporating $N_{sq}$ into training. However, this denoising ability may lack robustness due to the limited noise. To acquire a more extensive real-world noise set, we propose a patch-based negative augmentation to expand the noise domain.

\textit{Negative Augment toward Video Frames. }As illustrated in Fig.~\ref{fig:farmework} (b). We divide $V_{lr}$ into fixed-size patches. Negative augmentation will be applied in the patch-based scenario. Given the each patch sequence $V_{patch} \in  \mathbb{R}^{n \cdot s ^2    \times c \times \frac{h}{s} \times \frac{w}{s}} $. Meanwhile, $s$ represents the scale factor of the patches in $V_{lr}$. The high and width of each patch are $\frac{h}{s}\times \frac{w}{s}$. Expressed in the formula as: 

\begin{equation}
    V_{patch}=T(V_{lr}),
\end{equation}
where function $T(\cdot )$ denotes dividing $V_{lr}$ into $n \cdot s ^2$ patches of the same size, and the number of channels keeps constant. Then we apply negative augmentation to each patch $V_{patch}^i$.

\begin{figure}[t]
\centering
\includegraphics[width=1.0\linewidth]{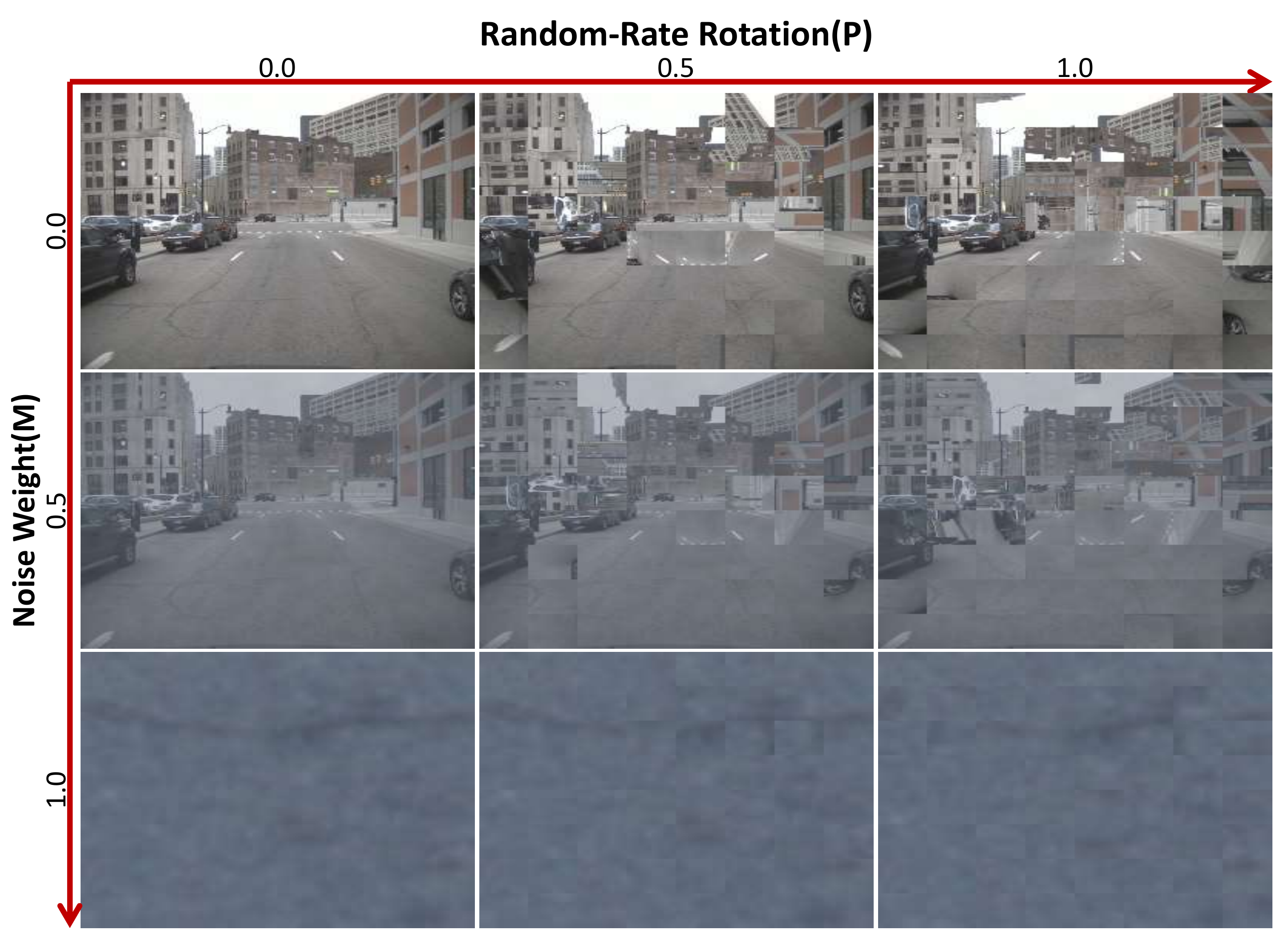}
\caption{A grid visualization of mixed images using the NegMix method by adjusting the noise weight (vertical) and rotation ratio (horizontal). We set $M$ to 0.5 and varied $P$ from 0 to 1 with an interval of 0.1 in our NegVSR setting. Zooming up for a better view.}
\label{fig:NegMix}
\end{figure}

\begin{figure}[t]
\centering
\includegraphics[width=0.98\linewidth]{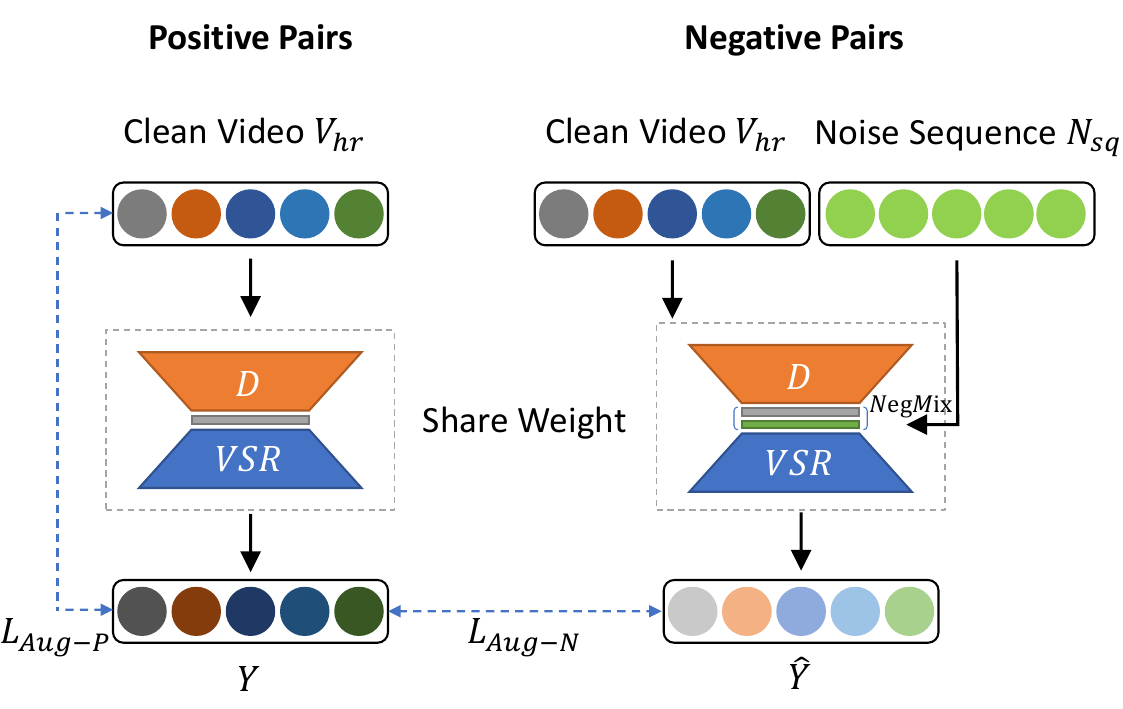}
\caption{The figure depicts the process of our Augmented Negative Guidance approach. We obtain the positive output $\widehat{Y}$ by passing $V_{hr}$ sequential through the degeneration model $D$ and VSR. Then we inject noise sequence $N_{sq}$ into the degraded video and apply the video with negative augmentation. Finally, we encourage the model to learn robust features from the augmented noise and video by $\mathcal{L}_{Aug-N}$ and $\mathcal{L}_{Aug-P}$. }
\label{fig:Loss}
\end{figure}

A random central rotation operation is performed on $V_{patch}^i$, with rotation angles of $[0, 90, 180, 270]$ degrees. For $V_{patch}$ under the same $V_{lr}$, patch-based rotation is applied with the same probability $P$. Each patch $V_{patch}$ is associated with corresponding practical rotation probability $p$. The probability $P$ is randomly selected from an array of $[0, 1]$ with an interval of 0.1. Likewise, $V_{patch}^i$ corresponds to a practical rotation probability $p$ randomly drawn from a uniform distribution [0, 1]. $Rot(\cdot)$ is only applied to the patch when $p$ is less than or equal to $P$. If $P$ equals 1, the $Rot(\cdot)$ is applied to all patches. It can be mathematically represented as:
\begin{equation}
Neg( V_{lr}, P)= \left \{
\begin{array}
{lr}Rot(T( V_{lr})),                    & p \leq P\\None, &otherwise
\end{array}\right.,
\end{equation}
where $Rot( \cdot  )$ refers to random central rotation operation and $None$ denotes without any augmentation. As illustrated in Fig.~\ref{fig:NegMix}, when $P$ approaches 1, less semantic information is preserved. Negative augmentation renders the semantic information unintelligible to the human. It poses a significant challenge to the capacity of VSR to reconstruct the information.

\textit{Negative Augment toward Noise Sequence. }$N_{sq}$ extracted from $V_{od}$ often consists of predominantly solid color blocks, which can negatively impact the generalization ability of VSR. To enhance the robustness of VSR to denoise ability, we also utilize negative augmentation for $N_{sq}$. Initially, we obtain $N_{sq}$ from $V_{od}$ using Sequential Real-world Noise Generation. $N_{sq}$ is then divided into patches, then a random central rotation operation is applied to each patch:
\begin{equation}
Neg( N_{sq}, P)= \left \{
\begin{array}
{lr}Rot(T( N_{sq})),                     & p \leq P\\None, &otherwise
\end{array}\right., 
\end{equation}
where $P$ should remain consistent for each pair of $V_{lr}$ and $N_{sq}$. The weight of $N_{sq}$ in the mixed sequence is controlled by $M$. Finally, our NegMix can be expressed as follows:
\begin{equation}
\begin{split}
    V_{neg}=NegMix(V_{lr}, N_{sq})=Neg(M \cdot N_{sq} \\+(1-M)\cdot V_{lr}, P).
\end{split}
\label{V_neg}
\end{equation}

\begin{figure*}[t]
\centering
\includegraphics[width=0.98\linewidth]{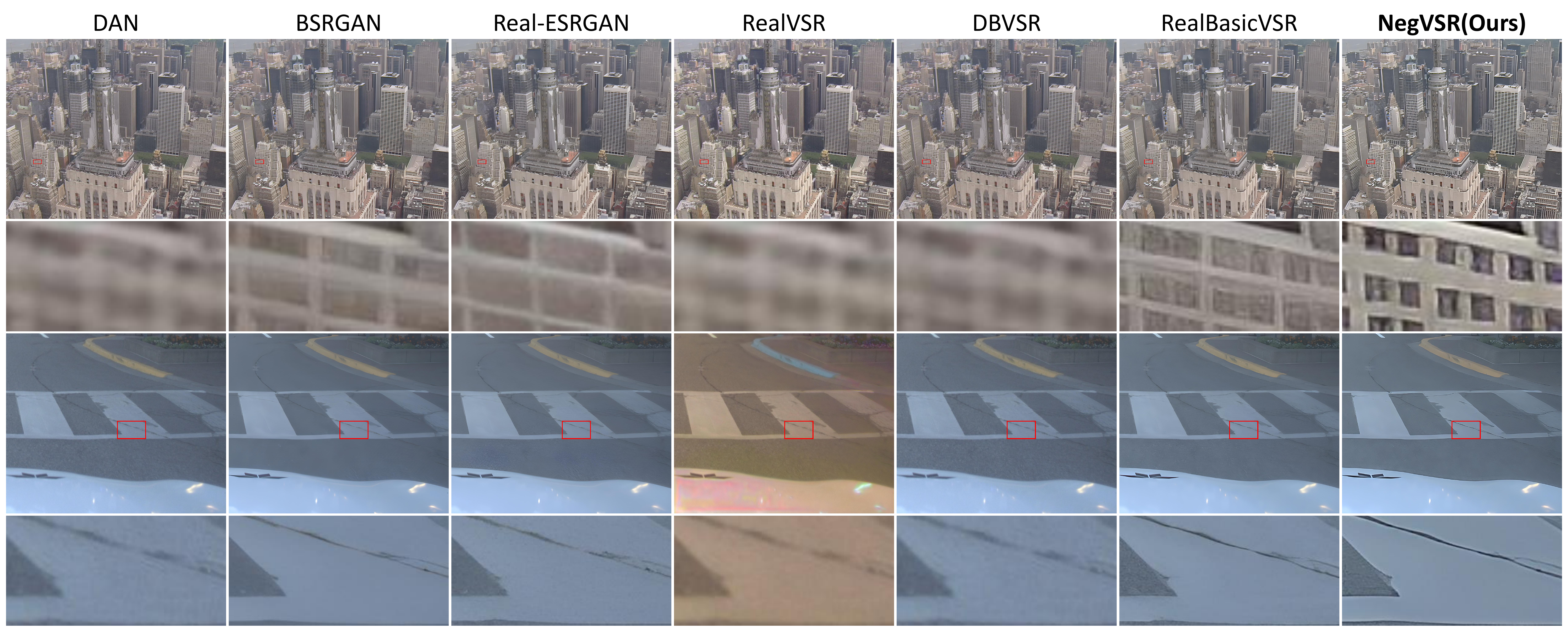}
\caption{We conduct a visual comparison with recent state-of-the-art methods on real-world images from the VideoLQ (1, 2 rows) and FLIR testing dataset (3, 4 rows), with the upsampling scale factor of 4.}
\label{fig:compartion}
\end{figure*}

\begin{algorithm}[tb]
\caption{NegVSR Training}
\label{alg:algorithm}
\textbf{Input}: HR video $V_{hr}$; Noise sequence $N_{sq}$;\\
\phantom{inputa:}Training iterations $M$;\\
\textbf{Output}: Final model $VSR_{M}$;
\begin{algorithmic}[1] 
\STATE Initialize the model $VSR_0$ randomly.
\FOR{$m \leftarrow 1$ \textbf{to} $M$}
\STATE Initialize model $VSR_m=VSR_{m-1}$;
\STATE Initialize the degeneration model $D$ randomly;
\STATE Degenerate Video $V_{lr}=D(V_{hr})$;
\STATE Generate negative video by Equ.~\ref{V_neg}:\\ $V_{neg}=NegMix(V_{lr}, N_{sq})$;
\STATE Calculate positive HR video by Equ.~\ref{pos output}: \\$Y=VSR_m(V_{lr})$;
\STATE Calculate negative HR video by Equ.~\ref{neg output}: \\$\widehat Y=VSR_m(V_{neg})$;
\STATE Calculate negative loss by Equ.~\ref{loss neg}: \\$\mathcal{L}_{Aug-N}(\widehat{Y}, Y)$;
\STATE Calculate positive loss by Equ.~\ref{eq:Loss pos}: \\$\mathcal L_{Aug-P}(V_{hr}, Y)$;
\STATE Optimize current model $VSR_m$ by minimizing the Equ.~\ref{eq:Loss_all}: $\mathcal L_{Aug-P}+\lambda\mathcal L_{Aug-N}$ for one iteration;
\ENDFOR
\end{algorithmic}
\end{algorithm}

\textbf{Recovering via Augmented Negative Guidance.} 
Given a clean video $V_{hr}$ and a degradation bank $D$~\cite{Real_ESRGAN}. The LR video $V_{lr}$ is degraded from HR video $V_{hr}$. We can apply the NegMix to $V_{lr}$ and then get negative output $\widehat{Y} $ through VSR. $Y$ represents the output of $V_{lr}$ via VSR.
\begin{equation}
    V_{lr}=D(V_{hr})=(D_n \cdot D_{n-1} \cdots D_2 \cdot D_1)(V_{hr}),
\label{D}
\end{equation}
\begin{equation}
    Y=VSR(V_{lr}),
\label{pos output}
\end{equation}
\begin{equation}
    \widehat{Y} =VSR(NegMix(V_{lr}, N_{sq})),
\label{neg output}
\end{equation}
where $D$ represents the degradation bank, which consists of various classic degradation kernels such as blur, resize, noise, and compression.

We propose an Augmented Negative Guidance that encourages the consistency between the augmented outputs (i.e., negatives $\widehat{Y} $ and positives $Y$). As shown in Fig.~\ref{fig:Loss}, we reconstruct the video without NegMix and only use the degradation bank $D$ to obtain the $Y$ through VSR. Next, we use NegMix on $V_{lr}$ to get $V_{neg}$, and then feed $V_{neg}$ into VSR to generate the corresponding negative output $\widehat{Y}$. Furthermore, our proposed approach minimizes the distance between the $V_{lr}$ prediction and its corresponding negative augmentation $V_{neg}$ output. It enables VSR to learn robust features in negative augmentation. We propose an Augmented Negative Guidance for  $\widehat{Y}$ as follows:

\begin{equation}
    \mathcal{L}_{Aug-N}(\widehat{Y} ^i, Y^i)=\frac{1}{B}\sum_{i=1}^B|| \widehat{Y} ^i, Y^i||_2,
\label{loss neg}
\end{equation}
where $B$ represents the batch size.

To promote the convergence of $V_{hr}$ and $Y$ as the positive augmented loss $\mathcal L_{Aug-P}$, various criteria (i.e., pixel loss, perceptual loss~\cite{perceptual_loss} and generative loss~\cite{gan_loss}) are utilized for Augmented Positive Guidance as:
\begin{equation}
\begin{split}
       \mathcal L_{Aug-P}(V_{hr}^i, Y^i)=\alpha \mathcal{L}_{Pix}(V_{hr}^i, Y^i)+ \\ \beta\mathcal{L}_{Per}(V_{hr}^i, Y^i)+\gamma\mathcal{L}_{Adv}(V_{hr}^i, Y^i),
\end{split}
\label{eq:Loss pos}
\end{equation}
where $\alpha = \beta =1.0$, $\gamma = 0.05$. 

$\mathcal{L}_{Aug-N}$ promotes performance and robustness by learning discriminative representations from augmented noise and frames. This regularization term can be seamlessly integrated into the loss function of VSR. By including this additional term, VSR is motivated to acquire characteristics resistant to negative augmentation, consequently advancing the generalization and recovering capacity. To this end, the total loss in our framework is summarized as follows:
\begin{equation}
    \mathcal{L}_{Aug-NP}=\mathcal{L}_{Aug-P}(Y, V_{hr})+\lambda \mathcal{L}_{Aug-N}(Y, \widehat{Y} ),
\label{eq:Loss_all}
\end{equation}
where $\lambda=0.5$ is the negative augmentation coefficient. 

\section{Experiment}
\subsection{Implementation Details}
\textbf{Training Setting. }We adopt the training setting of RealBasicVSR~\cite{RealBasicVSR} and train our NegVSR using the REDS~\cite{reds} dataset. Noise sequences are gathered from the FLIR training dataset ($V_{od}$). We employ the high-order degradation bank~\cite{Real_ESRGAN} to synthesize the training input. The size of $V_{hr}$ and $V_{lr}$ are $256\times256$ and $64\times64$, respectively. And we configure the patch size to be $4\times4$. Throughout the training process, the time length of the sequence $t$ fixes to 15. The flip inversion is employed to augment the sequence at each iteration. Setting batch size to 2. Optimizer adopts Adam. The SPyNet~\cite{spynet} model generates the optical flow estimation in the alignment module, and the SPyNet does not participate in the gradient backpropagation during the training process.

The training process comprises two distinct stages: pre-training and fine-tuning. During the pre-training phase, the model is trained for 100k iterations while maintaining the learning rate of $1\times10^{-4} $ and employing the $\mathcal{L}_{Aug-NP}$. The fine-tuning stage consisted of 150k iterations, where the learning rate is set to $5\times10^{-5}$. The loss function settings are consistent with Equ.~\ref{eq:Loss_all}. 

\textbf{Network Config. }We configure the propagation module ResBlock to 10 layers and set the ResBlock in the clean module to 20 layers. Additionally, the convolution kernel size is fixed at $3\times3$, and the number of middle channels is set to 64.


\begin{table*}[]
\small
\centering
\begin{tabular}{l|c|c|c|c|c|c|c|c|c}
                    & Bicubic & DAN    & BSRGAN & \begin{tabular}[c]{@{}c@{}}Real-\\ ESRGAN\end{tabular} & RealVSR & DBVSR  & \begin{tabular}[c]{@{}c@{}}RealBasicVSR,\\ our impl.\end{tabular} & \begin{tabular}[c]{@{}c@{}}RealBasicVSR,\\ original.\end{tabular} & NegVSR                     \\ \hline
NIQE $\downarrow$    & 7.987   & 7.086  & 4.204  & 4.187       & 7.810   & 6.732  & 3.936                                       & 3.699                                                            & \textbf{3.188}   \\
BRISQUE $\downarrow$ & 66.652  & 63.360 & 25.159 & 29.844      & 66.252  & 61.163 & 29.073                                       & 24.700                                                         & \textbf{22.255} \\
PI $\downarrow$      & 7.301   & 6.707  & 4.066  & 4.131       & 7.210   & 6.501  & 3.941                                        & 3.755                                                         & \textbf{3.416}  \\
NRQM $\uparrow$      & 3.392   & 3.740  & 6.155  & 6.053       & 3.432   & 3.796  & 6.195                                        & 6.313                                                             & \textbf{6.465}
\end{tabular}
\caption{The quantitative comparison of our proposed method with other VSR methods. Our method (NegVSR) exhibits superior performance compared to all other methods on the VideoLQ dataset. The metric is calculated on the Y channel. }
\label{tab:videoLQ}
\end{table*}

\begin{table*}
\centering
\small
\begin{tabular}{l|c|c|c|c|c|c|c|c|c}
                                                              & Bicubic & DAN                       & BSRGAN & \begin{tabular}[c]{@{}c@{}}Real-\\ ESRGAN\end{tabular} & RealVSR                  & DBVSR  & \begin{tabular}[c]{@{}c@{}}RealBasicVSR,\\ our impl.\end{tabular} & \begin{tabular}[c]{@{}c@{}}RealBasicVSR,\\ original.\end{tabular} & NegVSR                    \\ \hline
Params (M)                                                     & -       & 4.3  & 16.7   & 16.7        & 2.7   & 25.5   & 6.3                                                              & 6.3                                                               & 4.8                       \\
\begin{tabular}[c]{@{}l@{}}Runtimes\\ (ms/F)\end{tabular} & -       & 295.2 & 379.8  & 613.1       & 276.2 & 1280.2 & 387.5                                                            & 387.5                                                             & 315.0                     \\ \hline
NIQE $\downarrow$                                              & 8.656   & 8.253                     & 7.579  & 7.507       & 8.372                    & 8.348  & 6.464                                        & 6.096                                                            & \textbf{5.225}  \\
BRISQUE $\downarrow$                                           & 60.468  & 60.822                    & 32.396 & 37.905      & 59.398                   & 59.451 & 30.819                                       & 28.428                                                            & \textbf{20.702} \\
PI $\downarrow$                                                & 7.314   & 6.912                     & 5.508  & 5.681       & 7.153                    & 6.958  & 5.018                                       & 4.765                                                          & \textbf{4.201}  \\
NRQM $\uparrow$                                                & 3.915   & 4.383                     & 6.665  & 6.203       & 4.046                    & 4.380  & 6.695                                       & 6.829                                                               & \textbf{6.973}
\end{tabular}
\caption{Quantitative analysis of the FLIR testing dataset. The inference is performed on an NVIDIA 3090 24G with a fixed input frame size of $612\times512$, and the metric is calculated on the Y channel. }
\label{tab:flir}
\end{table*}

\subsection{Comparation} 
\textbf{Evaluation Dataset. }To comprehensively compare and validate our NegVSR, we employ the following two real-world VSR datasets, i.e., VideoLQ~\cite{RealBasicVSR} and FLIR testing dataset~\footnote{\url{https://www.flir.com/oem/adas/adas-dataset-form}}.

To keep consistent with the previous method~\cite{RealBasicVSR}, we calculate the image metrics for a portion of video included in both the VideoLQ and FLIR datasets to mitigate the computational overhead. Similarly, we select the first, middle, and last frames of each video. To FLIR frames, we divide the images into four equally sized copies to lower their resolution. Then the video is reorganized based on their segmented position. We select only the first 25 frames from each video.

\textbf{Evaluation Metris. }Due to the unavailability of labels, we conduct a quantitative assessment of reconstructed images using reference-free image quality assessment metrics such as NIQE~\cite{NIQE}, BRISQUE~\cite{BRISQUE}, NRQM~\cite{NRQM}, and PI~\cite{PI}.

\textbf{Evaluation Results. }We compare our approach with other VSR methods: DAN~\cite{DAN_1}, BSRGAN~\cite{BSRGAN}, Real-ESRGAN~\cite{Real_ESRGAN}, RealVSR\\~\cite{RealVSR}, DBVSR~\cite{DBVSR}, and RealBasicVSR~\cite{RealBasicVSR}. 'RealBasicVSR, original' refers to the RealBasicVSR officially released model. And 'RealBasicVSR, our impl' refers to the implementation of the RealBasicVSR with the same training settings as introduced in our paper.

The quantitative evaluation results of our experiments on VideoLQ are presented in Tab.~\ref{tab:videoLQ}. Our method exhibits superior performance in VideoLQ when compared to the other methods. Specifically, in contrast to RealBasicVSR, our method demonstrates a more effective blur removal. Fig.~\ref{fig:compartion} (1, 2 rows) exhibits the remarkable ability of NegVSR to remove blur and recover more details than other methods.

According to Tab.~\ref{tab:flir}, we demonstrate the metrics and runtimes test on the FLIR testing dataset, in which NegVSR achieves the best results among all evaluation metrics. A comprehensive depiction of the image details on FLIR is presented within Fig.~\ref{fig:compartion} (3, 4 rows). NegVSR shows a notably superior deblurring effect, enhancing the intricate texture of the road scene. And a satisfactory trade-off between computing speed and image quality is obtained.

\subsection{Ablations Study}
To evaluate the effectiveness of each component in NegVSR\\, we conducted an ablation comparison by separately analyzing each component. The baseline used in our ablation experiments represents RealBasicVSR. We performed a split on $\mathcal L_{Aug-NP}$. $\mathcal L_{Aug-P}$ indicates that only the loss function of Equ.~\ref{eq:Loss pos} is utilized. In contrast, $\mathcal L_{Aug-NP}$ indicates the usage of both $\mathcal L_{Aug-N}$ and $\mathcal L_{Aug-P}$. 'w/' indicates that we have incorporated additional components compared to the baseline. We employ VideoLQ as the test set.

\begin{table}[t]
\small
\centering
\begin{tabular}{l|c|c|c}
\textbf{Methods}         & $Loss$               & \textbf{NIQE} $\downarrow$  & \textbf{BRISQUE} $\downarrow$ \\ \hline
Baseline        & \multirow{4}{*}{$\mathcal L_{Aug-P}$} & 3.936 & 29.073  \\ \cline{1-1} \cline{3-4} 
w/ Noise         &                    & 3.643 & 25.286  \\ \cline{1-1} \cline{3-4} 
w/ Noise Sequences &                    & 3.215 & 22.951  \\ \cline{1-1} \cline{3-4} 
w/ NegMix        &                    & 3.312 & 22.969  \\ \hline
w/ NegMix        & $\mathcal L_{Aug-NP}$                  & \textbf{3.188} & \textbf{22.255 }
\end{tabular}
\caption{Ablation study of NegMix and $\mathcal L_{Aug-NP}$. Each proposed component is analyzed independently.}
\label{tab:ablation}
\end{table}

\textbf{Analysis of Noise Sequence. }In Tab.~\ref{tab:ablation}, the 'w/ Noise' denotes the noise mixed with the RealBasicVSR inputs during training. We employ the noise sampling method to extract from $V_{od}$. Specifically, $V_{od}$ is scanned using sliding windows of uniform size, and the noise is obtained by filtering these windows based on the calculation of their mean and variance. 'w/ Noise Sequences' utilizes our Sequential Real-world Noise Generation to extract $N_{sq}$ from the same $V_{od}$. The distribution of 'w/ Noise' is independent for each noise, and the noise domain of each 'w/ Noise Sequences' is identical. As shown in the Tab.~\ref{tab:ablation}, 'w/ Noise Sequences' outperforms both 'w/ Noise' and the baseline, suggesting that the proposed Sequential Real-world Noise Generation can effectively facilitate the utilization of this long-term noise in VSR.

\textbf{Recovering via Augmented Negative Guidance. }'w/ NegMix' refers to executing random center rotation for 'w/ Noise Sequences'. If 'w/ NegMix' is used without correcting the corrupted video with $\mathcal L_{Aug-N}$, the texture of the resulting image from 'w/ NegMix' will be distorted, leading to a degradation in performance as demonstrated in the Tab.~\ref{tab:ablation}. Utilizing NegMix with $\mathcal L_{Aug-NP}$ corresponds to our NegVSR. The closeness of positive and negative augmented outputs benefits VSR, enhancing its capacity to denoise robustly.

\section{Conclusion}
In this paper, we emphasized the significance of noise sequence in real-world VSR. In our study, we find that independent yet separate noise is not suitable for VSR tasks. Conversely, sequential noise exhibits a better solution in the VSR task. Despite efforts to address noise in real-world VSR, the monotonicity and finiteness of noise have resulted in many limitations, rendering the insufficient number for the task demands. To create more robust noise types for real-world VSR, we propose a Negatives augmentation strategy for generalized noise modeling. With the proposed NegVSR, the degeneration
domain is widely expanded by negative augmentation to build up various yet challenging real-world noise sets. We additionally present experiments on real-world datasets to show the effectiveness and superiority of NegVSR.

However, the proposed approach still has some limitations, especially the inference speed. In the following research, we are considering involving light-weight structures to facilitate real-time real-world VSR.

\section{Acknowledgements}
This work is supported in part by National Key R\&D Program of China (no.2021YFB2900900) and National Natural Science Foundation of China (NSFC) (no. 62002069).

\bibliography{aaai24}

\end{document}